\documentclass{article}
\pdfoutput=1

\usepackage{microtype}
\usepackage{graphicx}
\usepackage{subfigure}
\usepackage{booktabs} 

\usepackage{hyperref}

\usepackage[accepted]{icml2024}

\usepackage{amsmath}
\usepackage{amssymb}
\usepackage{mathtools}
\usepackage{amsthm}

\usepackage[capitalize,noabbrev]{cleveref}

\theoremstyle{plain}

\theoremstyle{definition}

\theoremstyle{remark}

\usepackage[textsize=tiny]{todonotes}

\usepackage{comment}
\usepackage{enumitem}
\setlist[itemize]{align=parleft,left=0pt..1em}   

\icmltitlerunning{ICML 2024 Workshop on LLMs and Cognition : Cryptic Verification}

\begin{document}


\twocolumn[
\icmltitle{Proving that Cryptic Crossword Clue Answers are Correct}



\icmlsetsymbol{equal}{*}

\begin{icmlauthorlist}
\icmlauthor{Martin Andrews}{rdai}
\icmlauthor{Sam Witteveen}{rdai}
\end{icmlauthorlist}

\icmlaffiliation{rdai}{Red Dragon AI, Singapore}

\icmlcorrespondingauthor{Martin Andrews}{martin@reddragon.ai}

\icmlkeywords{LLM,Cognition,ICML,Workshop}

\vskip 0.3in
]



\makeatletter\def\Hy@Warning#1{}\makeatother
\printAffiliationsAndNotice{}  


\begin{abstract}
Cryptic crossword clues are challenging cognitive tasks, 
for which new test sets are released on a daily basis by multiple international newspapers.  
Each cryptic clue contains both the definition of the answer 
to be placed in the crossword grid (in common with regular crosswords), 
and `wordplay' that \emph{proves} that the answer is correct 
(i.e. a human solver can be confident that an answer is correct without needing crossing words to confirm it).  
Using an existing cryptic wordplay proving framework 
(operating on Python proofs created by an LLM), 
we show that it is possible to distinguish between 
correct answers and almost-correct ones 
based upon whether the wordplay `works'.
\end{abstract}

\section{Introduction}
\label{introduction}

Recent advances in computational models have significantly improved 
their ability to handle diverse natural language tasks involving 
complex syntactic and semantic interpretations. 
Despite these strides, machines continue to fall short of human performance 
in areas requiring flexible problem-solving, swift adaptation to new tasks, 
and effective generalization across unfamiliar domains. 

This gap is particularly evident in the domain of cryptic crossword solving 
- a popular activity across the world, with multiple papers in
the UK, Australia, India and elsewhere featuring daily puzzles for readers to solve.

The domain of cryptic crossword solving has received little attention, 
despite being a notable language-oriented cognitive task, with solvers worldwide.
One possible reason is that cryptic crosswords are much less common in the United States
than `regular crosswords'.  
Another possibility is that cryptic crosswords combine a 
challenging cross-discipline mix of 
advanced language processing capabilities, 
logical reasoning, and an `Aha! moment'.  

The following illustrates the elements of a cryptic crossword clue 
(for more background please refer to the Appendix \ref{appendix-examples}):

{\footnotesize
\begin{verbatim}
clue:       Research done, primarily, 
            on most of magical beings (5)
definition: {Research} done, primarily, 
            on most of magical beings
wordplay:   D[one] (primarily) (most of) 
            ELVE[s] (magical beings)
answer:     DELVE
\end{verbatim}
}

In this example, the {\tt clue} is the text given to solvers 
(with the number of letters in the answer in brackets).
The reasoning steps include: 
(i) identifying the {\tt definition} (highlighted with curly braces), similar to a regular crossword; 
(ii) parsing the remainder of the clue to identify the key elements of the {\tt wordplay}.
Here, for instance, there are action words like `primarily' (meaning : take the first letter), 
and `most of' (meaning : remove some letters from) that are applied to other parts of the clue;
(iii) finally assembling the first letter of `done' and most of the letters in `elves' (the magical beings)
to `prove' that the correct {\tt answer} is `DELVE' (agreeing with the {\tt definition} span).

\begin{figure}
  \centering
{\footnotesize
\begin{verbatim}
# (1) Statement of original problem
def proof(
      clue="arrived with an artist, \
              to get optical device", 
      pattern="6", 
      answer="CAMERA"):  # Provided
  """
  # (2) Hypothesised by local LM
  definition: arrived with an artist, \
                to get {optical device}
  wordplay: CAME (arrived) + \
    RA (artist, from RA = Royal Academy)
  """
  # (3) Continuation generated by LLM 
  assert is_synonym("arrived", "CAME")
  assert is_abbreviation("artist", "RA")
  assert "CAME" + "RA" == "CAMERA"
  assert is_synonym(
     "optical device", "CAMERA",  
     pattern="6")
proof()   # Triggers proof verification
\end{verbatim}
}
\caption{From problem statement to LLM formalisation}
\label{WordplayFormalisation}
\end{figure}

In this work, 
taking a cue from the effectiveness of verifiers for other reasoning problems
\citep{lightman2023lets, jiang2023draft}, 
we approach the cryptic crossword clue solving problem as one that combines 
Language Models to tackle 
(i) the NLP elements; 
(ii) the creation an informal proof (i.e. coming up with {\tt wordplay}); 
(iii) the formalisation process (which re-writes the {\tt wordplay} logic in Python); and 
(iv) a `prover' that can check whether the claims are justifiable.
Rather than simply returning {\tt valid/invalid}, 
the prover provides `LLM-friendly' messages about validity, 
allowing the LLM to re-write its previous attempt iteratively.

\subsection{Contributions}
\label{Contributions}

The following is the main contribution of this work:

\begin{itemize}  
\item {\bf Show the effectiveness of the proving mechanism} - 
By using both the true {\tt answer} and a nearby {\tt candidate},
we show that the prover can distinguish between them based on the provability of the {\tt wordplay}
\end{itemize}

\section{Related Work}

\subsection{Regular Crosswords}
\label{Regular Crosswords}

Non-cryptic (``regular'') crosswords are known throughout the world, 
and are the predominant type found in newspapers in the U.S.A.
%
One key difference from cryptic crosswords is that regular crossword clues 
are generally not `standalone' - 
there may be a number of different answers that fit the given clue.  
The key to solving regular crosswords is thus the interaction between answers (i.e. the crossing-words), 
which allows for planning/backtracking to aid in breaking the combinatorial explosion of possibilities
to achieve solving rates in the high 90\% range \citep{wallace2022automated}.

\subsection{Cryptic Crosswords}
\label{Cryptic Crosswords}

In contrast to a regular crossword clue, a cryptic clue leads to its answer only if it is read in the right way. 
The clue itself contains both a conventional `straight definition', 
and wordplay that can be used to derive the same answer.
Once a given clue is understood, a solver can enter it into the grid with near 100\% certainty, 
even on a standalone basis.

To get a flavour of the mental processes involved in solving these puzzles, 
it is highly recommended to watch an expert going through the full process
for a recent Times Cryptic Crossword (including the reasoning steps in each clue)
\footnote{
Cracking the Cryptic (17-May-2024) \\
\url{https://youtu.be/vudt7LlUX00?t=124}
}.

Despite cryptic crosswords being being relatively unknown in the U.S.A, 
globally there are active communities of solvers, 
with multiple daily leaderboards and annual international competitions.

\subsection{Cryptonite Dataset}

The (UK) Times Cryptic Crossword is widely considered the gold standard in 
puzzles, even though they are not necessarily the most difficult, 
because the clues are unusually well constructed.
%
Cryptonite \citep{efrat2021cryptonite} 
is a large-scale dataset of Cryptic Crossword clues from the Times,
containing 523,000 naturally sourced clues from an extended time-period,
with the train, validation and testing splits chosen so that 
a given answer only appears in one of the splits.

\subsection{Rule-based solvers}
\label{Rule-based solvers}

\citet{williams1979computer} is an early example of attempting to devise 
a formal language for describing cryptic clues.
However, they found that the clues' linguistic elements tend to thwart such formal approaches.

\citet{rdeits-repo-python, rdeits-repo-julia} 
used a more flexible rule-based solver 
with a manually-crafted probabilistic grammar. 
Building on the assumption that a clue can usually be split into a wordplay and a definition, 
the (brute-force) solver tries to find the most probable parse such that the
wordplay yields a semantically-similar result to the definition.
Reported in \citet{efrat2021cryptonite}, the rule-based solver approach yields an accuracy of 8.6\%
on the Cryptonite test set.

\subsection{LLM-based solvers}
\label{LLM-based solvers}

Cryptic crossword clues seemed like an idea target for BERT-era models.
However, \citet{efrat2021cryptonite} reported that a T5-Large model fine-tuned 
on Cryptonite's 470k cryptic clue training set achieved only 7.6\% test set accuracy on the test set
(i.e. below that of rule-based solvers).

Interestingly, present day (scaled) Large Language Models also score very poorly on cryptic clues.
This is likely due to 
(i) the misleading surface reading of the clues; 
(ii) the obliqueness of the definitions; and 
(iii) the reasoning steps required to \emph{prove} the answer correct 
based on the wordplay that each clue provides.  


\subsection{Code \& reasoning}
\label{Code_reasoning}

To compensate for LLMs only approximating the generation of logical reasoning, 
techniques like PAL \citep{gao2023pal} exploit LLMs' facility 
for writing code to create verifiable reasoning chains.
An important influence on this work was also the Draft, Sketch, and Prove framework \citep{jiang2023draft} which 
uses an LLM to draft and create proofs that are then verified formally.

Informed by the evolution from AlphaCode \citep{Li_2022}, 
in which huge numbers of programs are generated and filtered in order to generate a valid solution,
to AlphaCodium \citep{ridnik2024code}, in which solutions are iterated upon
and involving much less computation, 
this work uses a prover that can feed back `hints' to the formalising LLM,
so that the task of re-writing nearly-valid proofs is made easier.

\section{Methods}
\label{Methods}

\subsection{Wordplay dataset}

There are a number of websites where cryptic crossword enthusiasts post 
completed puzzles, annotated with 
{\tt definition}, {\tt wordplay} and {\tt answer} fields.
In order to capture these key elements of cryptic crossword clue solving,
we make use of a Wordplay dataset gathered from such sites 
(further details in Appendix \ref{wordplay-dataset}).

\subsection{Language Model set-up}

In our experiments, we make use of two Language Models.

In order to generate the  {\tt definition} and {\tt wordplay} fields,
we make use of the Llama-3-it 8B model \citep{llama3modelcard}, fine-tuned using LoRA \citep{hu2021lora}
to generate {\tt definition} and {\tt wordplay} annotations from the original {\tt clue} and (importantly) a candidate {\tt answer}.
Training on 5371 examples (with the prompt format as shown in Appendix \ref{SFT-prompt}) 
took under 3 hours on a single GPU virtual machine, 
using the {\tt unsloth} package \citep{unsloth-repo}.

To create the python `proofs' of the correctness of solutions, 
we use both Google's {\tt Gemini-Pro-1.0-002} and {\tt Gemini-Flash-1.5-001} LLMs
(pinned model versions to enable a level of reproducibility).

While the Llama model was found to be capable of reasonable guesses at correct
{\tt definition} and {\tt wordplay} annotations, 
the creation (and iterative fixing) of the Python proofs required the use of more capable models.

\subsection{Hypothesis testing}

The hypothesis tested in this work is whether it is possible for the combination of
Llama {\tt definition} and {\tt wordplay} generation; Gemini LLM formalisation; and a Python-based prover 
to have sufficient `power' to distinguish between candidate answers 
(one of which is the correct answer).
Ideally, the {\it correct} answer will lead to perfect {\tt wordplay}, which then can be translated
into elegant Python code, while an {\it incorrect} candidate answer will lead to `bizarre' {\tt wordplay}, 
which in turn will be formalised into Python that will be incapable of being proved.

\begin{figure}
  \centering
{\footnotesize
\begin{verbatim}
def proof(answer="RUDE", 
  clue="rudeness about son’s computer language", 
  pattern='4'):
  """
  definition: 
    {rudeness} about son’s computer language
  wordplay: 
     RUD[e] (about, S (son)) + 
     ASS (assistant)
  """
  assert is_synonym("rudeness", 
    "RUDE", pattern='4') 
  assert is_abbreviation("son", "S")
  assert is_synonym(
    "assistant", "ASS") # Fails
  assert "RUD" + "ASS" == "RUDE" # Fails
proof()
# NB: correct answer is "LISP"
#   wordplay: 
#     (LIP) (rudeness) about (S) (son)
\end{verbatim}
}
\caption{Incorrect {\tt answer} leading to formalisation failure
}
\label{FailedWordplay}
\end{figure}

\subsection{Obtaining a close candidate answer}
\label{candidate-generation}

For a given question, we use the Llama model 
to create a {\tt definition} and {\tt wordplay} pair
from the {\tt clue} and the ground-truth {\tt answer}.  
We then use the span in the generated {\tt definition} to 
create an alternative candidate {\tt answer} 
that both matches the {\tt pattern} and is semantically
close to the phrase marked in the {\tt definition}.
This closest match is obtained by filtering a list of crossword words \citep{UKACD} 
sorted by cosine-similarity to the {\tt definition} span, 
when both are embedded using FastText \citep{FastText}.

\subsection{Formalising and proving an answer}
\label{formalisation-and-proving}

From a candidate {\tt answer}, we use the Llama model to generate 
a {\tt definition} and {\tt wordplay} pair.
We then use the Gemini LLM to attempt to generate Python proofs,
which are then verified using a Cryptic Crossword DSL expressed via Python
(see Appendix \ref{python-dsl} for further details).
This process includes `re-writes' where the proof verifier can 
return errors in response to assertion failures, along with hints about how these errors
might be fixed.  
After the initial draft proof, the verifier allows up to 5 re-write attempts to be made 
- until the proof is either accepted or the verification process stops 
(i.e. no success after 5 re-writes).

In the case of the close candidate answer,
the {\tt wordplay} is likely to be rather nonsensical - 
the hypothesis being tested here is whether the formalisation process can reject 
close candidate answers, in favour of the ground-truth answer.
Figure \ref{FailedWordplay} gives an illustration of the kind of output produced when a
non-ground-truth {\tt answer} is converted to {\tt wordplay} (and then an attempt at proving it is made).

\section{Experiments}

\subsection{Distinguishing ground-truth answers from close candidates}
\label{distinguishing}


For each of $100$ different clue examples from the Wordplay dateset,
we use the ground-truth answer to generate 1 close candidate answer, as in Section \ref{candidate-generation}.

We then provide the ground-truth and the candidate answers to Llama to 
generate $5$ different {\tt definition} and {\tt wordplay} samples for each.  

Given the {\tt definition} and {\tt wordplay}, we use the Gemini LLM 
to formalised the problem into Python 
(an example of which is shown in Figure \ref{WordplayFormalisation}), 
and then attempt verification of that Python proof, 
with a maximum of $5$ re-writes (attempts at re-formalisation) for each potential proof
(as in Section \ref{formalisation-and-proving}).

Finally, we gather the results (number of re-writes required for successful proof, or a fail) 
across all $100$ questions $\times$ 5 samples $\times$ 2 candidate answers.


To see whether the ground-truth answer was more `provable' than the close candidate,
we check which of them obtained:
(a) the higher number of completed proofs (of any number of re-writes);
(b) the fastest proof (i.e a proof requiring fewest re-writes); 
(c) the faster average solve time, where unsolved counts for $6$ re-writes (rather than infinity).

\section{Results}

The results of testing the `provability' hypothesis are 
shown in Table \ref{provability}, 
where we show percentages of 
True Positive (ground-truth answer more provable),
False Negative (non-ground-truth answer more provable) and
Draw (both answers proved to equal extents)
across the different provability measures, 
for each of the two Gemini models.


\begin{table}[t]
\caption{Frequency of `provability' wins by aggregation method \\
LLM version: 1.0P is Gemini-Pro-1.0, 1.5F is Gemini-Flash-1.5
}
\label{provability}
\vskip 0.15in
\begin{center}
\begin{small}
\begin{sc}
\begin{tabular}{lcccc}
\toprule
Method & LLM  & True & Draw & False \\
       & ver. & Pos  &      & Neg \\
\midrule
Completed Proofs & 1.0P & 38\% & 59\% &  3\% \\
Fastest Solve    & 1.0P & 38\% & 56\% &  6\% \\
Mean solve time  & 1.0P & 38\% & 56\% &  6\% \\
\midrule
Completed Proofs & 1.5F & 40\% & 55\% &  5\% \\
Fastest Solve    & 1.5F & 40\% & 55\% &  5\% \\
Mean solve time  & 1.5F & 42\% & 53\% &  5\% \\
\bottomrule
\end{tabular}
\end{sc}
\end{small}
\end{center}
\vskip -0.1in
\end{table}


Clearly, the results suggest that the proving system
has a degree of {\it preference} towards correct answers, 
but is a long way from being a reliable oracle of answer correctness.

This points to an issue that would likely occur if the system were scaled up
to testing many candidate answers, rather than just 2 possibilities here.
Specifically, if the cryptic crossword clue task were transformed to 
choosing between a large number of potential candidates
the current system would likely start to become less accurate overall, 
since the number of False Negative results would likely start to 
dominate the True Positive results.
That being said, there are many avenues for improvement, in particular
solving some of the limitations outlined in the Section \ref{Limitations}.

Looking across the LLM versions, 
it is also encouraging to see that the (much cheaper) Gemini-Flash model is 
slightly more capable of proving the ground-truth answers.

\section{Limitations}
\label{Limitations}

The Prover does not detect a number of potential errors / problems:

\begin{itemize}
\item Cryptic crossword setting `rules' dictate that the clues should contain exactly enough to prove an answer,
      the prover does not check that all valuable words in the clue have been utilised
\item Proofs may be logically disconnected, with left-hand-side terms 
      not necessarily being connected to right-hand-side terms in other lines of the code.  
\item Entire Python function consists of comments : Nothing triggers {\tt assert}
\item Python function contains conditional execution, routing around assert statements : Nothing triggers {\tt assert}
\item Occasionally, the hint {\tt assert XYZ failed} results in a re-write {\tt assert XYZ==False}, which is cheating
\end{itemize}

With additional effort, the authors believe that these issues are surmountable.
However, since the Gemini LLM is only being used In-Context,
there currently is little chance that the above issues are being systematically abused 
(which would almost certainly happen if there 
were learning-in-the-loop in a Reinforcement Learning setting).

\section{Conclusions}

It is increasingly hypothesised that the next-token-prediction task may be
insufficient to get machines to reason and plan \citep{Kambhampati_2024}.
By framing the cognitive task of cryptic crossword solving as a reasoning problem 
that is addressable by LLMs supported by a verification system, 
this work has sought to bring this reasoning task within the scope of
what is tractable by \emph{systems} that have components that include LLMs 
as well as verifiers and coding aids.

The authors sincerely hope that this work sparks an interest in the cryptic crossword domain, 
since it presents a challenging NLP/reasoning task, 
with huge scope for testing different reasoning approaches.
Notably, the current State-of-the-Art solving methods score less than 20\% on a real-world test set.

\newpage

\section*{Impact Statement}

There are many current cryptic crossword enthusiasts that would potentially not welcome
AI-enabled solvers to `take over' their favourite pastime.
In particular, when taken further, this line of work would be potentially disruptive
to public leaderboards that rank people according to the time taken to solve puzzles 100\% correctly.


However, there is currently little risk of LLM cryptic solvers as being anything more than
comic relief for current experts.  

Naturally, the authors also believe that the techniques here have wider
applications to the field of Machine Learning, 
but they do not in themselves present any particular additional societal risk.

\subsection*{Bias towards English-language speakers}
\label{Bias}

The English language has a high capacity for ambiguity and wordplay overall,
making cryptic crosswords much more feasible.  
However, they do exist in other languages - 
please see the Cryptic Crossword Wikipedia page for a broader view of their worldwide prevalence.
Note that deriving the answers is very difficult (even for native English speakers), 
whereas understanding the answer from given wordplay is much simpler.




\section*{Acknowledgements}

Support for this research was provided by the Google AI/ML Developer Programs team,
including access to the Gemini models
and GPUs on Google Cloud Platform. 

The authors thank 
the ICML workshop reviewers for their time and valuable feedback.

\bibliography{neurips_2024}
\bibliographystyle{icml2024}

\clearpage
\appendix
\onecolumn

\section{Cryptic Crossword Background}
\label{appendix-examples}

The following borrows extensively from the description on \citet{enwiki:1228427465} (kudos to the authors there), 
to which we have added {\tt wordplay} annotations in a notation typical of the \url{FifteenSquare.com} website 
(and in the Wordplay dataset use in this work).

\subsection{Basics}

A cryptic clue leads to its answer only if it is read in the right way. 
What the clue appears to say when read normally (the surface reading) is usually a distraction with nothing to do with the solution. 
The challenge is to find the way of reading the clue that leads to the solution.  

A typical clue consists of two parts:
\begin{itemize}
\item The straight or definition. 
This is in essence the same as any non-cryptic crossword clue: a synonym for the answer. 
It usually exactly matches the part of speech, tense, and number of the answer, and usually appears at the start or end of a clue.  
For our annotations, the span that encompasses the {\tt definition} is highlighted using curly braces.
\item The cryptic, subsidiary indication or wordplay. 
This gives the solver some instructions on how to get to the answer in another (less literal) way. 
The wordplay parts of clues can be obscure, especially to a newcomer, 
but they tend to utilise standard rules and conventions which become more familiar with practice.
\end{itemize}

Sometimes the two parts of the clue are joined with a link word or phrase such as `from', `gives' or `could be'. 
One of the tasks of the solver is to find the boundary between the definition and the wordplay, 
and insert a mental pause there when reading the clue cryptically.

We list below several of the important styles of {\tt wordplay} that are commonly used, 
each with an annotated example.  
For a more comprehensive list, along with an outline of the `Ximenean principles', please see \citet{enwiki:1228427465}.

\subsection{Anagrams}

An anagram is a rearrangement of a certain section of the clue to form the answer.
This is usually indicated by a codeword which indicates change, movement, breakage or something otherwise amiss.
For example:

%
%

{\footnotesize
\begin{verbatim}
clue:       Chaperone shredded corset (6)
definition: {Chaperone} shredded corset
answer:     ESCORT
wordplay:   (corset)* (*shredded)
\end{verbatim}
}

\subsection{Charade}

In a charade, the answer is formed by joining individually clued words to make a larger word (namely, the answer).  
For example:

{\footnotesize
\begin{verbatim}
clue:       Outlaw leader managing money (7)
definition: Outlaw leader {managing money}
answer:     BANKING
wordplay:   BAN (outlaw) + KING (leader)
\end{verbatim}
}

\subsection{Containers}

A container or insertion clue puts one set of letters inside another.
For example (also starting to add a little more indirection):

{\footnotesize
\begin{verbatim}
clue:       Utter nothing when there's wickedness about (5)
definition: {utter} nothing when there's wickedness about
answer:     VOICE
wordplay:   O (nothing) with VICE (wickedness) around it (about)
\end{verbatim}
}

\begin{samepage}
\subsection{Deletions}

Deletion is a wordplay mechanism which removes some letters of a word to create a shorter word.
For example:
\nopagebreak
{\footnotesize
\begin{verbatim}
clue:       Bird is cowardly, about to fly away (5)
definition: {Bird} is cowardly, about to fly away
answer:     RAVEN
wordplay:   [c]RAVEN (cowardly) - 'C' (i.e. circa, about) (-fly away)
\end{verbatim}
}
\end{samepage}

\subsection{Double definition}

A clue may, rather than having a definition part and a wordplay part, have two definition parts.
For example:

{\footnotesize
\begin{verbatim}
clue:       Not seeing window covering (5)
definition: {Not seeing} {window covering}
answer:     BLIND
wordplay:   Double Definition (DD)
\end{verbatim}
}

\subsection{Hidden words}

With hidden word clues, the solution itself is written within the clue – either as part of a longer word or across more than one word.
For example:

{\footnotesize
\begin{verbatim}
clue:       Found ermine, deer hides damaged (10)
definition: Found ermine, deer hides {damaged}
answer:     UNDERMINED
wordplay:   [fo]UND ERMINE D[eer] (hides)
\end{verbatim}
}

\subsection{Homophones}

Homophones are words that sound the same but have different meanings, such as `night' and `knight'. 
Homophone clues always have an indicator word or phrase that has to do with being spoken or heard.
For example:

{\footnotesize
\begin{verbatim}
clue:       We hear twins shave (4)
definition: We hear twins {shave}
answer:     PARE
wordplay:   "pair" (twins, "we hear")
\end{verbatim}
}

\subsection{Reversals}

A word that gets turned around to make another is a reversal.
For example:

{\footnotesize
\begin{verbatim}
clue:       Returned beer fit for a king (5)
definition: Returned beer {fit for a king}
answer:     REGAL
wordplay:   (LAGER)< (beer, <returned)
\end{verbatim}
}

\newpage
\section{Wordplay Dataset}
\label{wordplay-dataset}

The Wordplay Dataset used in this work is extracted from websites 
where cryptic crossword enthusiasts post solutions
to the puzzles published in major publications.  
Each completed puzzle is annotated by an solver
who provides the community with {\tt definition}, {\tt wordplay} and {\tt answer} fields
for each of the approximately 30 clues in that day's grid.  

For UK papers, these enthusiast websites include:
\begin{itemize}
\item \url{timesforthetimes.co.uk} -  Times, Times Quick 
\item \url{www.fifteensquared.net} -  Independent, Guardian, Financial Times
\item \url{bigdave44.com} -  Telegraph, Sunday Telegraph
\end{itemize}

\begin{samepage}
The following is an example from the Wordplay dataset, formatted in YAML:

{\footnotesize
\begin{verbatim}
title: Financial Times 16,479 by FALCON
url: https://www.fifteensquared.net/2020/05/18/ \
     financial-times-16479-by-falcon/
author: teacow
clues:
- clue: '{Offer} of support also broadcast'
  pattern: '8'
  ad: D
  answer: PROPOSAL
  wordplay: PROP (support) + (ALSO)* (*broadcast)
- ...
\end{verbatim}
}
\nopagebreak
In the above:
\begin{itemize}
\item {\tt clue} is the original clue, as given to solvers, 
but with the `regular crossword' {\tt definition} portion highlighted with curly braces; 
\item {\tt pattern} is the number of characters in the answer;
\item {\tt ad} (across/down) is potentially significant, because some clues include
directional hints such as `before' or `upwards' which are only meaningful if the orientation 
of the answer within the grid is known; 
\item {\tt answer} is the clue's final answer (not known to the solvers before solving); 
and
\item {\tt wordplay} is an informally annotated explanation of how the clue words act together
to logically build the letters in the answer 
(the resulting grid letters typically being in upper case)
- here the {\tt *} symbol signifies that {\tt ALSO} is to be anagrammed 
due to the anagram indicator ({\tt broadcast}) in the clue.
\end{itemize}
\end{samepage}

Code that generates the Wordplay dataset is available at \url{https://github.com/mdda/cryptic-wordplay}.
Note that care has been taken to ensure that 
the training/validation/test splits follow those of the Cryptonite dataset
(and the test set is deliberately not provided, to reduce the chance that it becomes training data itself).

\newpage
\section{Fine-tuning prompt}
\label{SFT-prompt}

The following is a verbatim training example used for the fine-tuning of the Llama-3-it model:

{\footnotesize
\begin{verbatim}
<|start_header_id|>system<|end_header_id|>

Cryptic clue wordplay generation : Given the clue and the answer, \
return expert definition and wordplay annotations<|eot_id|>\
<|start_header_id|>user<|end_header_id|>

clue: "musical and ballet, oddly, that can be avoided"
answer: EVITABLE ~ evitable<|eot_id|>
<|start_header_id|>assistant<|end_header_id|>

definition: musical and ballet, oddly, {that can be avoided}
wordplay: EVITA (musical) + B[a]L[l]E[t] (ballet, odd letters)<|eot_id|>\
<|end_of_text|>
\end{verbatim}
}

\section{In-Context Learning Prompts for the Gemini LLM}
\label{python-dsl}

The Gemini LLM is prompted in-context with the concatenation of the following sections: 
\begin{itemize}
\item Cryptic Crossword overview
\item Many-shot wordplay examples
\item Declaration of `external' Python functions
\item 6-shot formalisation demonstration
\item Actual problem statement (for continuation as a Python proof)
\item {\it After a verification failure}: Error messages for the generated proof, with hints if available, and request to improve iteratively
\end{itemize}

The sections of the prompt are described more fully below, 
note that care was taken to ensure that the chosen terminology was use consistently throughout.

\subsection{Cryptic Crossword preamble}

The following is the rubric and {\tt wordplay} preamble given to the Gemini LLM:

{\footnotesize
\begin{verbatim}
A Cryptic crossword question involves using the words in \
the given clue to yield an answer that matches the letter pattern.  
The clue will provide a definition of the answer, as well \
as some 'wordplay' that can also be used to confirm the answer.  
Expert question solvers write informal 'proofs' using a \
particular format.

For the definition, the original clue is annotated with \
'{}' to denote where the definition is to be found.
For the wordplay, the following conventions are loosely used:
* The answer is assembled from the letters in CAPS
* Words in brackets show the origins of letters in CAPS, \
often being synonyms, or short forms 
* Action words are annotated as illustrated:
  + (ETO N)* (*mad = anagram-signifier) = TONE
  + (FO OR)< (<back = reversal-signifier) = ROOF
  + [re]USE (missing = removal-signifier) = USE
* DD is a shorthand for 'Double Definition'
\end{verbatim}
}

\subsection{Many-shot wordplay examples}

Around 20 examples from the Wordplay dataset are included in the in-context prompt:

{\footnotesize
\begin{verbatim}
For example:
---
clue: "arrived with an artist, to get optical device (6)"
definition: arrived with an artist, to get {optical device}
answer: CAMERA
wordplay: CAME (arrived) + RA (artist, short form)
---
clue: ...
\end{verbatim}
}

\subsection{External Python DSL functions}

Domain Specific Python functions are described in-context to the LLM,
which appears able to use them without seeing their internal functionality.
In fact, the actual implementation of the functions is more extensive than
described, since calls to these functions also track `near misses' which 
can be fed back as hints during the re-write process.

{\footnotesize
\begin{verbatim}
The task is to produce a formal proof using python code, \
where the docstring will also include an informal proof as an aid.
The following are functions that can be used in your output code:

Action=Enum('Action', 'ANAGRAM,REMOVE_FIRST,INITIALS,REMOVE_LAST,'+
                      'GOES_INSIDE,GOES_OUTSIDE,REVERSE,SUBSTRING,HOMOPHONE')
# External definitions
def is_synonym(phrase:str, test_synonym:str, pattern:str='') -> bool:
  # Determines whether 'test_synonym' is a reasonable synonym for 'phrase', 
  # with letters optionally matching 'pattern'
def is_abbreviation(phrase:str, test_abbreviation:str) -> bool:
  # Determines whether 'test_abbreviation' is 
  # a valid abbreviation or short form for 'phrase'
def action_type(phrase:str, action:Action) -> bool:
  # Determines whether 'phrase' might signify the given 'action'
def is_anagram(letters:str, word:str) -> bool:
  # Determines whether 'word' can be formed from 'letters' (i.e. an anagram)
def is_homophone(phrase:str, test_homophone:str) -> bool:
  # Determines whether 'test_homophone' sounds like 'phrase'
\end{verbatim}
}

\subsection{Few-shot formalisation examples}

The following are 3 (out of 6) of the few-shot formalisation examples given
before the final test-case prompt:

{\footnotesize
\begin{verbatim}
The following are examples of simple functions that prove that \
each puzzle solution is correct:

```python
def proof(answer="ONCE", 
          clue="head decapitated long ago", pattern='4'):
  """
  definition: head decapitated {long ago}
  wordplay: [b]ONCE (head decapitated = remove first letter of BONCE) 
  """
  assert is_synonym("head", "BONCE")
  assert action_type("decapitated", Action.REMOVE_FIRST) \
         and "BONCE"[1:]=="ONCE"
  assert is_synonym("long ago", "ONCE", pattern='4')
proof()
```

```python
def proof(answer="DECIMAL", 
          clue="the point of medical treatment", pattern='7'):
  """
  definition: {the point} of medical treatment
  wordplay: (MEDICAL)* (*treatment = anagram) 
  """
  assert is_synonym("the point", "DECIMAL", pattern='7')
  assert action_type("treatment", Action.ANAGRAM)
  assert is_anagram("MEDICAL", "DECIMAL")
proof()
```

```python
def proof(answer="SUPERMARKET", 
          clue="fat bags for every brand that’s a big seller", 
          pattern='11'):
  """
  definition: fat bags for every brand that’s {a big seller}
  wordplay: SUET (fat) (bags = goes outside) of \ 
            (PER (for every) + MARK (brand)) 
  """
  assert is_synomym("fat", "SUET")
  assert action_type("bags", Action.IS_OUTSIDE)
  assert "SUET" == "SU" + "ET"
  assert is_abbreviation("for every", "PER")
  assert is_synomym("brand", "MARK")
  assert "SU"+"PER"+"MARK"+"ET" == "SUPERMARKET"
  assert is_synonym("a big seller", "SUPERMARKET", pattern='11')
proof()
```
\end{verbatim}
}

\subsection{Formalisation instruction}

The following instruction is given before the final `test-case' prompt illustrated in Figure \ref{WordplayFormalisation}:

{\footnotesize
\begin{verbatim}
# Please complete the following in a similar manner, and return the whole function:

```python
def proof(answer= ...
\end{verbatim}
}

\subsection{Proof Verification with Hinting}
\label{Proof-Verification-with-Hinting}

Examples of assertion failures, with constructive hinting, are shown:

{\footnotesize
\begin{verbatim}
AssertionError: assert: is_abbreviation('an Artist', 'RA') : 
   'an Artist' does not have a valid abbreviation; 
   'RA' is an abbreviation for : artist, artillery, Royal Artillery, 
   gunners, painter
AssertionError: assert action_type('goes crazy', Action.ANAGRAM) : 
  'goes crazy' itself does not suggest Action.ANAGRAM, but 'crazy' does
AssertionError: assert action_type('worked', Action.HOMOPHONE) : 
  'worked' does not suggest Action.HOMOPHONE, but maybe Action.ANAGRAM  

# Please re-implement the SOLUTION above \
(altering both the docstring and the python code as required), \
taking care to fix each of the problems identified, \
and return the whole function:

```python
def proof(answer= ...
\end{verbatim}
}

Once the prover has fully parsed a given output with zero assertion failures, 
the proof is considered a success 
(up to 5 re-write iterations are allowed, more that that is considered an overall failure to prove the answer).

%
%

\end{document}